\documentclass{article}
\usepackage{arxiv}
\newcommand{\papertitle}{Building a hierarchical exchange into a graph neural network for power flow}
\newcommand{\paperkeywords}{power flow, graph neural networks, Kron reduction, graph coarsening, grid foundation models}
\newcommand{\paperauthors}{Carmine Delle Femine, Leire Garin Atxaga, Asier Diaz-Iglesias, Juan Pablo Maroto Herrera, Ane Miren Florez-Tapia, Marco Quartulli, Izaro Goienetxea Urkizu}

\newcommand{\tablefont}{\small}
\usepackage[T1]{fontenc}
\usepackage[utf8]{inputenc}
\usepackage{amsmath,amssymb}
\usepackage{graphicx,booktabs}
\usepackage{xcolor,tikz}
\usetikzlibrary{arrows.meta}
\usepackage{microtype}
\usepackage[font=small,labelfont=bf]{caption}
\usepackage{placeins}
\usepackage{natbib}
\setcitestyle{authoryear,round,aysep={},yysep={,},citesep={;}}
\usepackage{url}
\usepackage[colorlinks=true,allcolors=teal]{hyperref}
\hypersetup{
  pdftitle={\papertitle},
  pdfauthor={\paperauthors},
  pdfkeywords={\paperkeywords}
}
\usepackage{orcidlink}
\renewcommand{\shorttitle}{\papertitle}
\renewcommand{\undertitle}{preprint}
\title{\papertitle}
\author{\normalfont
  \begin{minipage}{0.97\textwidth}
  \centering
  Carmine Delle Femine\textsuperscript{1,2}\,\orcidlink{0009-0005-7964-8911}, Leire Garin Atxaga\textsuperscript{1}\,\orcidlink{0009-0002-4194-4940}, Asier Diaz-Iglesias\textsuperscript{1}\,\orcidlink{0009-0003-1156-3191},\\
  Juan Pablo Maroto Herrera\textsuperscript{1}\,\orcidlink{0009-0006-1532-4413}, Ane Miren Florez-Tapia\textsuperscript{1}\,\orcidlink{0000-0002-7259-5731},\\
  Marco Quartulli\textsuperscript{1}\,\orcidlink{0000-0001-5735-2072}, Izaro Goienetxea Urkizu\textsuperscript{2}\,\orcidlink{0000-0002-1959-131X}\\[5pt]
  {\small \textsuperscript{1}Vicomtech Foundation, Basque Research and Technology Alliance (BRTA), Department of Energy,\\
  20009 Donostia/San Sebasti\'an, Spain\\
  \textsuperscript{2}University of the Basque Country (UPV/EHU), Department of Languages and Information Systems,\\
  20018 Donostia/San Sebasti\'an, Spain\\
  \href{mailto:cdellefemine@vicomtech.org}{cdellefemine@vicomtech.org}}
  \end{minipage}
}
\date{}

\begin{document}
\maketitle
\begin{abstract}
Solving AC power flow couples every bus to the whole network, whereas corrective graph neural networks exchange information only between neighboring buses. We insert a two-level hierarchical exchange into GENCO, replacing two of its twelve corrective layers at lower counted matrix FLOPs. The exchange compresses bus states onto nested sets of anchor buses, mixes them by coarse-graph message passing and global attention, and maps the context back to every bus. Bus-to-anchor weights come from Kron's zero-injection map of the admittance matrix (Kron) or a graph partition (Quotient). Under a 200-epoch multi-grid training across \mbox{500-,} \mbox{2,000-,} and 4,917-bus grids, Kron lowers the flat network's error on new scenarios by 85.0\% (from 3.24 to 0.49 times a bus-mean reference) and Quotient by 78.2\%. Both beat the reference on every training grid and seed, with Kron consistently ahead, while no model beats it on two unseen grids.

\end{abstract}

\section{Introduction}

Operating and planning an electric power grid requires solving the AC power-flow problem repeatedly to determine the complex voltage at every bus from the power injected at each bus. Classical iterative solvers \citep{wardhale1956}, accelerated by ordered sparse triangular factorization \citep{tinneyhart1967,tinney1967direct} and active--reactive decoupling \citep{stottalsac1974}, remain the standard, but evaluating many operating scenarios can become costly, motivating learned surrogates based on graph neural networks (GNNs). A central challenge for such surrogates is that, although each nodal balance equation involves only a bus and its immediate neighbors, the resulting voltages depend on the entire network. The voltages follow from the injected currents through the inverse of the sparse admittance matrix, which is dense \citep{tinney1967direct}: for example, on the largest grid studied here (4,917 buses), 0.07\% of the entries of the admittance matrix are nonzero, while no entry of its inverse vanishes. Moreover, perturbations in AC power flow can induce slow-decaying, long-range responses across transmission topologies \citep{jung2016longrange}.

Graph neural networks are natural candidates for learning this dependence, because they operate directly on graphs \citep{scarselli2009gnn}. In the message-passing form, each node holds a state updated over successive steps from its neighbors' messages \citep{gilmer2017mpnn}. A power grid is such a graph, with buses as nodes and branches as edges. Because the learned updates are shared across buses, one trained model can run on grids of any size, although its accuracy on a new grid must be measured.

Early power-flow GNNs were trained through Kirchhoff-law residuals \citep{donon2020graph}. Later models introduce higher-order graph convolutions on transmission grids \citep{lin2024powerflownet}, physics-based losses across distribution-grid topologies \citep{bottcher2023realistic}, decentralized message passing \citep{hansen2023decentralized}, and modularity-guided community structure \citep{xu2026modularity}, while grid foundation models seek representations shared across operating conditions, tasks, and networks \citep{gridfm}. Corrective networks approach the solution through repeated message passing between neighboring buses, decoding provisional voltages and feeding nodal mismatches back into the bus states. GENCO \citep{genco}, the corrective neural solver of the GridFM development framework\footnote{Implementation: \url{https://github.com/gridfm/gridfm-graphkit}.}, stacks twelve such layers.

Message passing is local by construction. A node can receive information from another bus a given number of branch hops away, measured as the shortest-path distance in transmission branches (edges) of the grid graph, only after at least that many exchanges between neighbors \citep{alon2021bottleneck}. On large grids, buses can lie beyond a model's reach, the largest hop distance across which one bus can influence another. The hop diameter, defined as the largest shortest-path distance in branch hops between any two buses, is 23, 41, and 50 on the three benchmark training grids used in this study (500, 2,000, and 4,917 buses; Section~\ref{sec-setup}), respectively. GENCO's twelve corrective layers each perform one round of message passing, and all but the last add a second exchange by feeding back nodal mismatches computed from neighboring voltages. A decoded voltage can then depend only on buses within 23 hops. On the 4,917-bus grid, for instance, 40\% of the bus pairs lie farther apart than that. GENCO's authors name the locality bottleneck of message passing as a limitation and leave scalable long-range mechanisms to future work \citep{genco}.

Adding layers to close this gap can carry a price. Graph convolutions can oversmooth with depth, as node states grow similar \citep{li2018deeper,rusch2023survey}, and messages from distant nodes can be oversquashed into fixed-size vectors \citep{alon2021bottleneck,topping2022curvature}. The evidence on power grids is mixed. One power-flow study names oversmoothing as the likely cause of the weaker accuracy of a deep convolution stack \citep{hansen2023decentralized}. A state-estimation study observes no oversmoothing in its best deep models \citep{ringsquandl2021power}, and PowerFlowNet reaches its minimum loss at most buses within about three hops \citep{lin2024powerflownet}.

A coarse level offers another route, because it shortens the paths between distant buses without adding layers. In power-system analysis, reduced equivalents predate digital solvers \citep{ward1949} and keep an equivalent network among retained buses, exact when no current is injected at the eliminated ones \citep{kron} and minimizing dissipation in resistive circuits \citep{vanderschaft2010}; multiscale graph-signal transforms likewise combine node selection, Kron reduction, and interpolation \citep{shuman2016}. Multigrid methods address an analogous limitation of local iterations: because each update inspects only immediate neighbors, local steps quickly damp sharp error differences between adjacent nodes, whereas slowly varying long-range errors appear nearly constant locally and decay very slowly; transferring them to a coarser level brings distant regions closer together so they can be corrected efficiently \citep{brandt1977,briggs2000}.

In learned simulators, multiscale GNNs build hierarchies over meshes to reach higher accuracy at equal step count \citep{fortunato2022multiscale} or enable longer-range interactions with fewer iterations \citep{bistride}, as well as over domain decompositions \citep{mggnn}. Graph coarsening by multilevel partitioning \citep{metis}, cell aggregation with spectral and cut guarantees \citep{loukas2019}, and learnable pooling layers \citep{diffpool,gao2019graphunets} construct related hierarchies for graph- and node-level tasks. These precedents do not predict the effect of a coarse level on a learned power-flow network.

GENCO has no coarse level, so we replace two of its corrective layers with a \emph{hierarchical exchange}. The exchange compresses the bus states onto retained buses, the \emph{anchors}, passes information among them on a coarse graph, and returns the result to every bus. For illustration, on the 4,917-bus grid, 984 anchors split the remaining buses into 2,037 groups of at most 26 buses, so every path between two groups passes through an anchor. Extending reach motivates this design, although our experiments compare architectures overall rather than isolating reach as the cause of the error reduction (Section~\ref{sec-conclusion}). We test whether this exchange improves GENCO's corrective network when a single model is shared across three grids on new operating scenarios, and whether the coarse graph should follow the electrical equations or a graph partition.

Both hierarchical models, Kron and Quotient, use the same anchors and differ in how each bus distributes its weight over them. Kron takes each bus's weights from the zero-injection map of the admittance matrix, which eliminates the other buses electrically \citep{kron}. Quotient takes each bus's weights from a graph partition, which puts all of a bus's weight on the anchor of its cell \citep{metis,loukas2019}. In both models, a fixed geometry computed per grid carries learned updates shared across grids inside GENCO's corrective backbone; everything else, training included, is shared with the flat network (Flat), GENCO's unmodified Base architecture.

In a 200-epoch multi-grid training, the exchange lowers the error of the flat network on new scenarios of the training grids by 85.0\% with Kron weights and by 78.2\% with Quotient weights. On every training grid in every seed, both hierarchical models beat the bus-mean reference (which predicts each bus's mean training voltage), whereas the flat network remains above it; because our training is far shorter than IBM's training of GENCO, this comparison isolates architectural differences rather than measuring GENCO as published (Section~\ref{sec-genco-relation}). On two held-out unseen grids excluded from training, no model beats the reference.

Section~\ref{sec-method} describes the exchange, Section~\ref{sec-setup} the training and evaluation, Section~\ref{sec-results} the results, and Section~\ref{sec-conclusion} the conclusions and limitations.

\section{Hierarchical exchange}
\label{sec-method}

\subsection{Backbone}

All models apply GENCO's corrective layers to a heterogeneous input graph of buses, generators, and branches. Each layer exchanges features over this graph, decodes provisional voltages and generator quantities, computes branch flows and nodal mismatches, and feeds the mismatch back into the bus states. Known quantities are kept by projection.

Flat is the unmodified GENCO Base architecture \citep{genco}, with twelve corrective layers. The hierarchical models have ten and apply the exchange after layer five. The exchange updates bus states, and generator states pass through it unchanged.

All corrective layers use hidden size 48 and eight attention heads, so a bus state $z$ entering the exchange has 384 channels. Flat has 20,069,187 parameters, and each hierarchical model has 19,882,947.

\subsection{One reduction level}

\begin{figure}[!htbp]
\centering
\includegraphics[width=\linewidth]{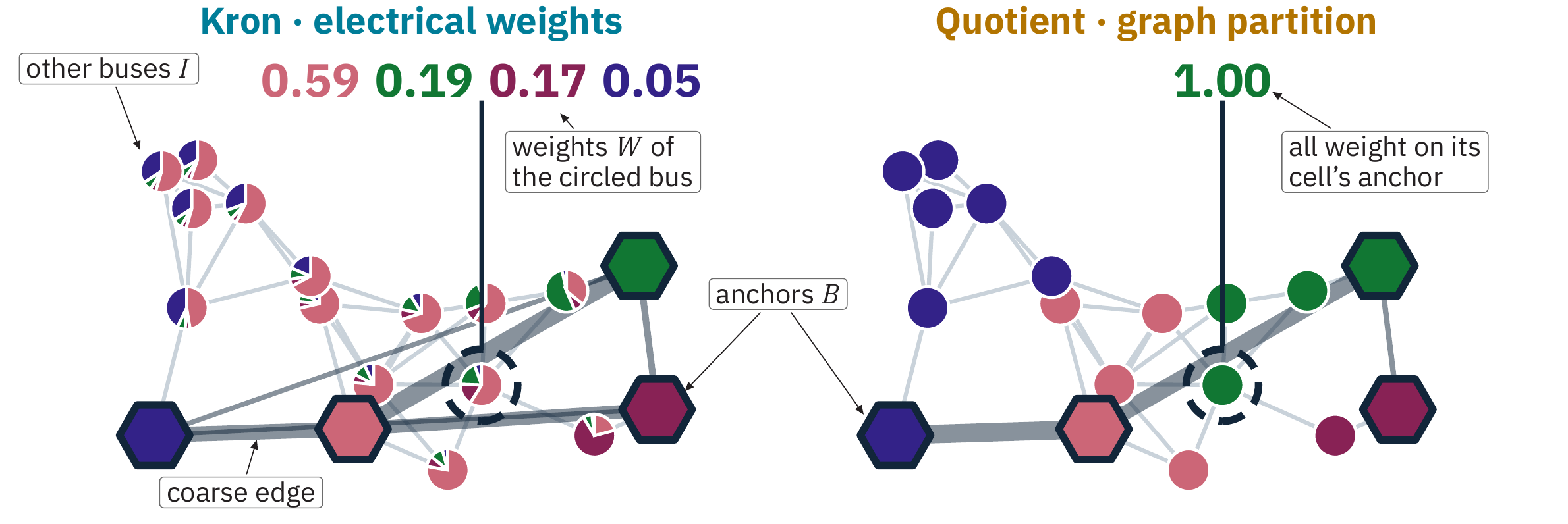}
\caption{Kron and Quotient weights on an illustrative 18-bus graph with admittance $Y$ and the same four anchors (hexagons) for both. Each pie shows a bus's weights $W$ as shares, its row of $P$, spread over several anchors for Kron and all on its cell's anchor for Quotient. The numbers give the circled bus's shares. Coarse edges, thicker when stronger, carry the Kron-reduced admittance or the admittance summed between cells. A toy example, not grid data.}
\label{fig-coarsening}
\end{figure}

At one level, let $B$ be the $m$ retained buses, the \emph{anchors}, and $I$ the other $N-m$ buses, ordered $(B,I)$, and let $Y\in\mathbb C^{N\times N}$ be the nodal admittance matrix. At the first level, METIS \citep{metis} partitions the buses into cells, and each cell's anchor is its bus of highest degree. Kron and Quotient use the same anchors and describe each bus of $I$ by a row of nonnegative weights over the anchors, collected in $W\in\mathbb R^{(N-m)\times m}$. They differ in how $W$ is chosen (Figure~\ref{fig-coarsening}).

\textbf{Kron.} Partitioning the nodal current--voltage equations $i=Yv$ into anchor and non-anchor blocks yields $i_B=Y_{BB}v_B+Y_{BI}v_I$ and $i_I=Y_{IB}v_B+Y_{II}v_I$. Setting the injected currents at the non-anchor buses to zero ($i_I=0$) gives
\begin{equation}
v_I=Kv_B,\qquad K=-Y_{II}^{-1}Y_{IB},
\label{eq-zero-injection}
\end{equation}
where $K\in\mathbb C^{(N-m)\times m}$ is the zero-injection map: entry $K_{ij}$ expresses how the voltage at non-anchor bus $i\in I$ responds to the voltage at anchor $j\in B$ when no current is injected at $I$. Substituting $v_I=Kv_B$ into the anchor equation yields $i_B=Sv_B$, where $S=Y_{BB}-Y_{BI}Y_{II}^{-1}Y_{IB}$ is the Schur complement, or Kron-reduced admittance \citep{kron}. Matrix $S$ joins anchors connected through eliminated buses, supplies the coarse-graph couplings for message passing among anchors, and serves as the reduced admittance matrix for the second reduction level.

Kron's bus-to-anchor weights are the magnitudes $W=|\widetilde K|$ of a truncation $\widetilde K$ of $K$. To keep the first-level transports sparse rather than scaling as $(N-m)m$, $\widetilde K$ retains at most the sixteen largest magnitudes in each row of $K$; on the training grids, the kept entries carry at least 99.6\% of each row's total magnitude. Both $K$ and $S$ are computed once per grid by solving $Y_{II}K=-Y_{IB}$ through lower--upper (LU) triangular factorization of $Y_{II}$, without explicitly forming $Y_{II}^{-1}$.

\textbf{Quotient.} Quotient's binary weights come from a partition of all buses into cells, one per anchor, with $W_{ij}=1$ if bus $i$ is in the cell of anchor $j$ and 0 otherwise. In $(B,I)$ order, the membership matrix of all buses is $H=[\mathrm{Id}_m;\,W]\in\{0,1\}^{N\times m}$, where $\mathrm{Id}_m$ is the $m\times m$ identity matrix mapping each anchor to its own cell. The anchors are coupled by the coarse admittance $Q=H^{\mathsf T}YH$, which sums the admittance between cells \citep{loukas2019} and plays the same role as $S$ for the coarse graph.

\textbf{Coarse edges and transports.} The coarse graph is directed. Each anchor sends to the anchors of the eight largest off-diagonal magnitudes above $10^{-3}$ per unit in its row of $S$ or $Q$ and receives from every anchor that selects it. Edge attributes are the entry's real part, imaginary part, magnitude, and phase.

Because $W$ collects unnormalized complex magnitudes in Kron and variable cell sizes in Quotient, its rows and columns need not sum to one; normalizing them turns state transport into scale-preserving weighted averages. With row sums $a=W\mathbf 1_m$ and column sums $b=W^{\mathsf T}\mathbf 1_{N-m}$, prolongation and restriction are
\begin{equation}
P=\operatorname{diag}(a)^{-1}W,\qquad R=\operatorname{diag}(b)^{\dagger}W^{\mathsf T},
\label{eq-transport}
\end{equation}
where $\dagger$ denotes the pseudoinverse, which inverts nonzero diagonal entries and leaves zero entries unchanged (in case an anchor receives no weight from $I$). In both variants, prolongation $P$ assigns each bus of $I$ the $W$-weighted mean of the anchor states, and restriction $R$ assigns each anchor the $W$-weighted mean of the states of the non-anchor buses mapped to it. For Kron, these means are weighted by the zero-injection magnitudes $|\widetilde K|$; for Quotient, where $W$ is binary, $P$ copies the cell's anchor state and $R$ takes the unweighted average of the other buses in the cell.

\subsection{The exchange}
\label{sec-exchange}

\begin{figure}[!htbp]
\centering
\includegraphics[width=\linewidth]{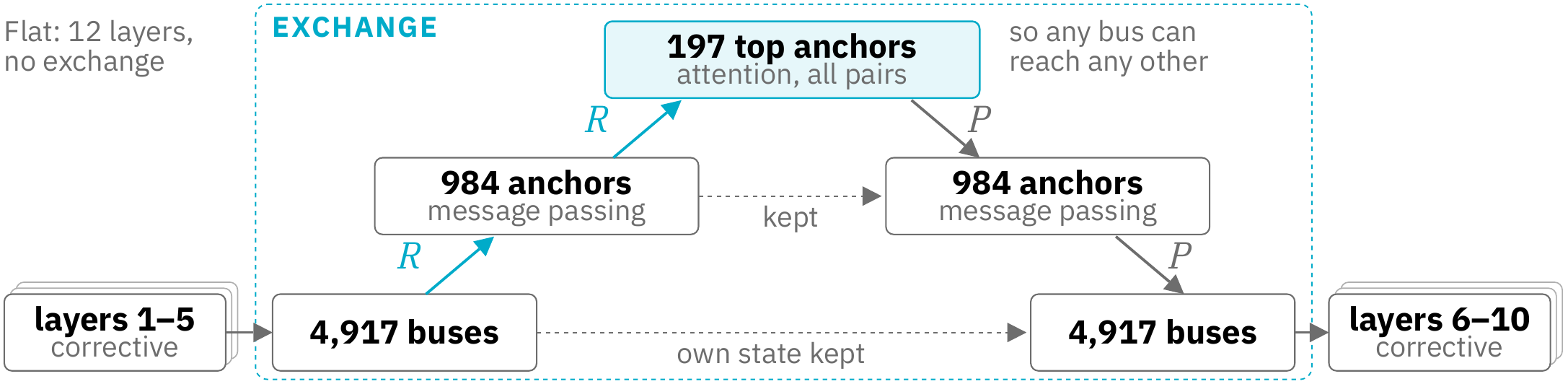}
\caption{The exchange, illustrated with the node counts of the 4,917-bus grid. Flat has twelve corrective layers, while the hierarchical models have ten and apply the exchange after layer five. Restriction $R$ carries bus states up to the first-level anchors (984 on this grid) and top anchors (197), and prolongation $P$ returns them to every bus. With attention among all pairs of top anchors, the state of any bus can depend on that of any other. Dashed arrows carry the states kept for the way down. Anchors, $W$ (at most 16 anchors per bus at the first level), $R$, $P$, and coarse edges (at most 8 leaving each anchor) are fixed per grid. The ten layers, $\phi_R$, $\phi_P$, message passing, and attention, about 20 million parameters as in Flat, are learned once for all grids.}
\label{fig-exchange}
\end{figure}

Two successive reductions each retain one fifth of the nodes (rounded up), with the second applied to the anchors of the first. Across the three training grids, this yields 100, 400, and 984 first-level anchors and 20, 80, and 197 \emph{top anchors} on the \mbox{500-,} \mbox{2,000-,} and 4,917-bus grids, respectively, which is few enough for all-pairs attention at the top level (illustrated for the 4,917-bus grid in Figure~\ref{fig-exchange}). To choose the top anchors, which are shared by both hierarchical models, we first form an unweighted graph on the first-level anchors keeping only strong electrical couplings, defined as off-diagonal entries of $S$ whose magnitude is at least one tenth of the largest off-diagonal entry in their row. Because $S$ is denser than $Y$, this row-relative threshold filters out weak fill-in entries while keeping every anchor connected to its dominant couplings. We then select the top anchors greedily by farthest-point sampling on this graph: starting from one anchor, each step adds the anchor at largest shortest-path hop distance from those already chosen, spreading the top anchors evenly across the reduced grid. Because $S$ is already small (for example, $984\times 984$ on the 4,917-bus grid), the second Kron level applies the same reduction to $S$ without truncating $K$; the second Quotient level assigns each anchor to its nearest top anchor in hops on the unweighted first quotient graph.

Since $R$ and $P$ are fixed, two in-place rules with learned MLPs $\phi_R$ and $\phi_P$, shared between levels, move information up and down:
\begin{equation}
z_B\leftarrow z_B+\phi_R(Rz_I),\qquad
z_I\leftarrow z_I+\phi_P([z_I,\,Pz_B]),
\label{eq-rules}
\end{equation}
where $[\cdot,\cdot]$ is concatenation and, at each level, $z_B$ and $z_I$ are the states of the anchors and the other buses. In restriction, each anchor adds a learned update of the pooled context $Rz_I$ to its own state before coarse-graph exchange; in prolongation, $\phi_P$ receives both the bus's local state $z_I$ and the returned coarse context $Pz_B$ so that the update can depend on how the returned context compares with the local state.

On the way up, restriction updates the first-level anchors from the non-anchor buses mapped to them. Message passing then runs among these anchors on the coarse graph, and a second restriction carries their states to the top anchors. Attention among all pairs of top anchors, with four heads, is the only step that mixes every pair.

On the way down, anchors take the states returned by the coarser level, and the remaining buses apply the prolongation rule to their states kept on the way up. A second message-passing block runs among the first-level anchors before the last prolongation. Because every bus has positive weight on at least one anchor at each level, every bus state can depend on every other bus state after the exchange.

Anchors, weights, transports, and coarse edges are fixed for each grid and computed from its topology and admittance. The ten corrective layers, $\phi_R$, $\phi_P$, and the message-passing and attention blocks are learned once for all grids.

\section{Training and evaluation}
\label{sec-setup}

\subsection{Grids and training}
\label{sec-genco-relation}

Each model is trained jointly on three GOC benchmark grids from PGLib\footnote{\url{https://github.com/power-grid-lib/pglib-opf}.} with 500, 2,000, and 4,917 buses (the \emph{training grids}), while two held-out grids with 3,022 and 4,020 buses are excluded from training (the \emph{unseen grids}). Operating scenarios vary the loads on each grid's intact topology, with 1,864 training scenarios for the 500- and 2,000-bus grids, 1,664 for the 4,917-bus grid, and 217 validation scenarios per grid. Each epoch draws 1,856 scenarios per training grid in single-grid batches of sixteen by cycling through reshuffled permutations of the grid's split, so some scenarios of the 4,917-bus grid appear twice per epoch. Across the three grids, this yields 348 updates per epoch and 69,600 updates over 200 epochs (23,200 per grid).

All three architectures share the same training policy and are trained with three random seeds each; we evaluate the final checkpoints of all nine completed runs. Optimization uses AdamW in full precision with $(\beta_1,\beta_2)=(0.9,0.999)$ and an initial learning rate of $5\times10^{-4}$, held constant for the first twenty epochs and then multiplied by 0.7 by a plateau scheduler whenever validation loss fails to improve for five consecutive epochs. Input normalization is fitted once per training grid.

The training objective combines two losses from GENCO: a supervised voltage error and an intermediate-layer power-balance penalty, weighted as
\begin{equation}
\mathcal L_t=0.9\,\mathcal L_{\mathrm{bus}}+0.1\min\!\left(1,\frac{t}{6960}\right)\mathcal L_{\mathrm{phys}},
\label{eq-loss}
\end{equation}
where $t$ counts gradient updates. For a scenario of $N$ buses (averaged over each single-grid batch), the two terms are
\begin{equation}
\mathcal L_{\mathrm{bus}}=\frac{1}{|\mathcal M|}\sum_{(i,c)\in\mathcal M}(\hat y_{i,c}-y_{i,c})^2,\qquad
\mathcal L_{\mathrm{phys}}=\sum_{\ell=1}^{L} w_\ell\,\frac{1}{N}\sum_{i=1}^{N}\sqrt{\bigl(\Delta P_i^{(\ell)}\bigr)^2+\bigl(\Delta Q_i^{(\ell)}\bigr)^2},
\label{eq-loss-terms}
\end{equation}
where $\mathcal M$ indexes the unknown bus voltage entries (magnitude in per unit and phase angle in radians), with final predicted and ground-truth values $\hat y_{i,c}$ and $y_{i,c}$. In $\mathcal L_{\mathrm{phys}}$, $\Delta P_i^{(\ell)}$ and $\Delta Q_i^{(\ell)}$ are the active and reactive nodal power-balance mismatches at bus $i$ after corrective layer $\ell$ (in normalized per-unit power), and $w_\ell=0.5^{L-\ell}/\sum_{k=1}^{L}0.5^{L-k}$ are normalized geometric weights over the $L$ corrective layers ($L=12$ for Flat, $L=10$ for Kron and Quotient) that favor later layers. The physics weight ramps up over the first twenty epochs, whereas validation always uses the full-strength objective.

All three models inherit their loss formulation, loss weights, normalizer, optimizer, and plateau factor from IBM's released configuration of GENCO Base \citep{genco}, while our training budget is far shorter and simpler than IBM's. Per grid, our 23,200 updates are 13 times fewer than the 300,600 updates of IBM's released Small GENCO model for the 2,000-bus grid\footnote{\url{https://huggingface.co/gridfm/genco-pf-datakit}.} (trained on about 16,000 scenarios), and IBM's released Base model for the 500-bus grid was configured for 200 epochs on 175,834 training scenarios (94 times our 1,864 scenarios for that grid), with N-$k$ topology perturbations that our intact-topology splits omit. Accordingly, these experiments compare architectures under a controlled training budget rather than measuring GENCO as published.

\subsection{Evaluation}

Test scenarios are generated with fresh random seeds to form pools of 2,331 scenarios per grid, from which 200 scenarios per grid (chosen by identifier before generation) form shared test panels: 600 scenarios across the training grids and 400 across the unseen grids. Each grid requires a per-grid input normalizer (GENCO's scalar power-scaling factor, computed from a percentile of loads and solved generator powers) as well as its bus-mean reference. On the training grids, both are fitted on the training splits. On each unseen grid, the model weights are never updated, but because this normalizer requires solved scenarios, both the normalizer and the reference are fitted on a separate split of solved calibration scenarios (disjoint from the test panel); we therefore refer to unseen-grid evaluation as \emph{calibrated transfer} rather than strictly zero-shot transfer.

Because voltage magnitude (in per unit) and phase angle (in degrees) have different physical units, we combine them into a single dimensionless \emph{combined voltage error} $e$ by normalizing each component's RMSE by a fixed reference scale and weighting the two equally:
\begin{equation}
e=\sqrt{\frac12\left[\left(\frac{r_V}{0.01\,\mathrm{p.u.}}\right)^2+\left(\frac{r_\theta}{1^\circ}\right)^2\right]},
\label{eq-error}
\end{equation}
where $r_V$ is the voltage-magnitude RMSE and $r_\theta$ is the wrapped angle RMSE over the unknown entries. We aggregate scenario errors within each grid and average equally across the three training grids to obtain the \emph{macro error}, reporting its mean and sample standard deviation over the three seeds.

As a simple baseline, the \emph{bus-mean reference} predicts each bus's mean solved voltage over the training split (or calibration split on an unseen grid), independent of the scenario. Alongside raw macro errors and their ratio to this reference, we report the share of test scenarios on which a model's seed-averaged error falls below the reference's error on the same scenario.

\section{Results}
\label{sec-results}

\subsection{Training}

\begin{figure}[!htbp]
\centering
\includegraphics[width=\linewidth]{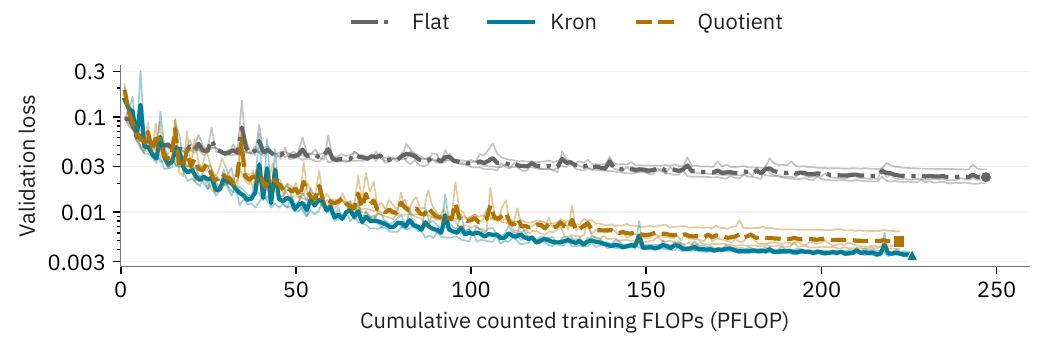}
\caption{Validation loss of the nine trainings against cumulative counted training FLOPs, on a logarithmic loss scale ($1~\mathrm{PFLOP}=10^{15}$~FLOP). Thin curves show individual seeds, bold curves the three-seed mean, and end markers each model's last epoch. Each run has 200 epochs and 69,600 updates (13 times fewer updates per grid than IBM's released GENCO model for the 2,000-bus grid). Validation uses the objective of Eq.~\eqref{eq-loss} at full physics weight, not the test error of Eq.~\eqref{eq-error}.}
\label{fig-training}
\end{figure}

Under the same training policy, both hierarchical models converge to substantially lower validation losses than Flat (Figure~\ref{fig-training}), with the three-seed mean curves of Kron and Quotient both finishing below Flat's.

To compare computational cost alongside convergence, we estimate training FLOPs using PyTorch's FLOP counter for the forward and backward matrix operations, adding the exchange's sparse transports, measured on two batches per model and grid and scaled to the 116 batches per grid and epoch. This count excludes pointwise, reduction, and scatter operations, optimizer updates, validation, and preprocessing, and therefore measures neither wall-clock runtime nor hardware cost. Within this scope, a 200-epoch run requires 246.94~PFLOP for Flat, 225.88~PFLOP for Kron, and 222.12~PFLOP for Quotient: replacing two corrective layers with the hierarchical exchange reduces counted matrix FLOPs by 8.5\% for Kron and 10.1\% for Quotient.

\subsection{New scenarios on the training grids}

On new operating scenarios of the training grids, adding the hierarchical exchange lowers macro error dramatically relative to Flat: by 85.0\% with Kron weights (84.3 to 85.5\% per seed) and by 78.2\% with Quotient weights (74.3 to 81.3\% per seed). Across the 600 test scenarios, the macro error in Table~\ref{tab-results} drops from $5.660\pm0.899$ for Flat to $0.851\pm0.110$ for Kron and $1.235\pm0.225$ for Quotient, compared with 1.747 for the bus-mean reference.

\begin{figure}[!tb]
\centering
\includegraphics[width=\linewidth]{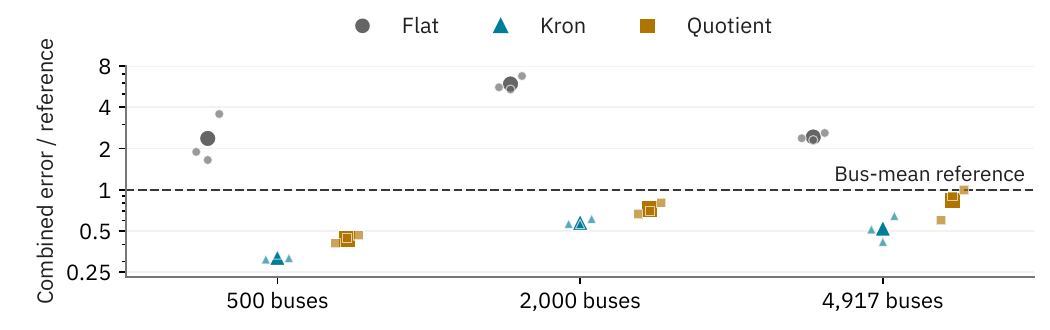}
\caption{Combined voltage error (Eq.~\eqref{eq-error}) relative to each training grid's bus-mean reference, on 200 new scenarios per grid. Small markers show individual seeds and large markers show three-seed means; values below one outperform the bus-mean reference.}
\label{fig-reference}
\end{figure}

Relative to that reference, Kron achieves 0.49 times the reference error and Quotient 0.71 times, whereas Flat stays at 3.24 times (2.4, 5.9, and 2.4 times on the \mbox{500-,} \mbox{2,000-,} and 4,917-bus grids, respectively; Section~\ref{sec-genco-relation}). As Figure~\ref{fig-reference} shows, every seed of both hierarchical models lies below the reference on every training grid, while Flat stays above it; the narrowest margin is Quotient on the 4,917-bus grid in one seed, 0.4\% below the reference (2.675 against 2.686). Evaluated scenario by scenario with seed-averaged errors, Kron outperforms the reference on 98.5\% of the 600 scenarios, Quotient on 91.7\%, and Flat on 1.3\%.

Comparing the two hierarchical constructions isolates the effect of electrical weighting, since Kron and Quotient share identical anchors, architecture, and parameter count. Under this training, Kron outperforms Quotient on every training grid in every seed, with Quotient's macro error ranging from 1.2 to 1.7 times Kron's across seeds. Distributing each bus's state across its electrically coupled anchors via the admittance matrix therefore yields a consistent accuracy advantage over assigning each bus solely to its partition cell's anchor.

\begin{table}[!tb]
\centering
% Values: results/2-multi-grid-data-volume/1x-data/metrics_by_run_and_grid.csv (fresh-panel rows) and
% poster/figs/w24r/numbers.json (bus-mean references, shares below the reference); recomputed 21 October 2026.
{\tablefont
\setlength{\tabcolsep}{6pt}
\renewcommand{\arraystretch}{1.16}
\begin{tabular}{lrrrr}
\toprule
\textbf{Grid} & \textbf{Bus mean} & \textbf{Flat} & \textbf{Kron} & \textbf{Quotient} \\
\midrule
\multicolumn{5}{l}{\emph{Training grids}}\\
500 buses & 1.310 & 3.104 \ensuremath{\pm} 1.372 & \textbf{0.421 \ensuremath{\pm} 0.017} & 0.574 \ensuremath{\pm} 0.038 \\
2,000 buses & 1.245 & 7.361 \ensuremath{\pm} 0.919 & \textbf{0.722 \ensuremath{\pm} 0.037} & 0.900 \ensuremath{\pm} 0.091 \\
4,917 buses & 2.686 & 6.516 \ensuremath{\pm} 0.408 & \textbf{1.411 \ensuremath{\pm} 0.307} & 2.230 \ensuremath{\pm} 0.558 \\
Macro & 1.747 & 5.660 \ensuremath{\pm} 0.899 & \textbf{0.851 \ensuremath{\pm} 0.110} & 1.235 \ensuremath{\pm} 0.225 \\
Macro / bus mean & 1 & 3.24 & \textbf{0.49} & 0.71 \\
Scenarios below bus mean & --- & 1.3\% & \textbf{98.5\%} & 91.7\% \\
\midrule
\multicolumn{5}{l}{\emph{Unseen grids, calibrated transfer}}\\
3,022 buses & \textbf{1.333} & 7.304 \ensuremath{\pm} 0.555 & 5.389 \ensuremath{\pm} 0.261 & 7.754 \ensuremath{\pm} 1.507 \\
4,020 buses & \textbf{0.880} & 13.910 \ensuremath{\pm} 0.768 & 12.576 \ensuremath{\pm} 0.975 & 12.190 \ensuremath{\pm} 0.966 \\
\bottomrule
\end{tabular}%
}%

\caption{Combined voltage error (Eq.~\eqref{eq-error}) on the test panels of new scenarios (200 per grid), reported as mean $\pm$ sample standard deviation over three seeds. The macro row weights the three training grids equally, with its standard deviation taken over the three seed macros. The bus-mean reference is fitted on the training split (or on the calibration split for unseen grids), and the share of scenarios below the bus mean uses seed-averaged errors. Best value in each row is in bold (lower is better, except for the share of scenarios below the bus mean).}
\label{tab-results}
\end{table}

\subsection{Unseen grids}

In calibrated transfer to the two unseen grids (3,022 and 4,020 buses), no model beats the bus-mean reference in any seed (Table~\ref{tab-results}). Errors remain several times above the reference for all three architectures. For example, Kron reaches 5.389 against a reference of 1.333 on the 3,022-bus grid and 12.576 against 0.880 on the 4,020-bus grid, with Flat and Quotient showing comparable gaps. Although the hierarchical exchange adapts its per-grid geometry to new topologies, learning shared weights from only three training grids is not yet sufficient for generalization to unseen network structures.

\FloatBarrier
\section{Conclusion and next steps}
\label{sec-conclusion}

Replacing two of GENCO's twelve corrective layers with a hierarchical exchange substantially improves multi-grid power-flow accuracy under a 200-epoch training on three grids, while reducing counted matrix FLOPs. On new scenarios of the training grids, the exchange lowers macro error relative to the unmodified flat network by 85.0\% with Kron weights (from 3.24 to 0.49 times the bus-mean reference) and by 78.2\% with Quotient weights, bringing both hierarchical models below the reference on every training grid and seed. Because Kron and Quotient share the same anchors and parameter budget, Kron's consistent advantage over Quotient demonstrates that electrical weighting from the admittance matrix outperforms purely topological cell aggregation at this data volume. On two held-out unseen grids, however, no model beats the reference in calibrated transfer, pointing to cross-topology generalization and stronger physical consistency as the main next steps.

Several limitations bound these conclusions. The training budget is short (at most 1,864 scenarios and 23,200 updates per grid, with load variations on intact topologies and three seeds per model), so the results compare architectures under a shared budget rather than evaluating GENCO as trained by IBM, and whether Kron's advantage over Quotient persists at larger data volumes remains open. Finally, while comparing Kron and Quotient isolates the electrical weights and coarse couplings, separating the individual contributions of reach, anchor placement, coarse message passing, and top-level attention (noting that the 500-bus grid already has a 23-hop diameter within GENCO's reach) is left to a more extensive study.

\section*{Acknowledgments}
% Funding statement and the disclaimer prescribed by the Horizon Europe MGA, art. 17.3.
\textbf{Funding.} This work was carried out in the context of SYSTEMICO, co-funded by the European Union under grant agreement No 101269929. Views and opinions expressed are however those of the authors only and do not necessarily reflect those of the European Union or the European Climate, Infrastructure and Environment Executive Agency (CINEA). Neither the European Union nor CINEA can be held responsible for them. The authors thank the GridFM community for the open-source framework.

\textbf{Use of AI-assisted tools.} Large language models (specifically Claude Opus 5.5 and Gemini 4 Neon) were used to assist with code debugging, prose polishing, and manuscript revision. All experimental designs, implementations, numerical results, and the final text were reviewed, verified, and approved by the authors, who take full responsibility for the content of this article.

\begingroup
\small
\setlength{\bibsep}{2pt}
\bibliographystyle{chicago}
\bibliography{article}
\endgroup
\end{document}